\pdfoutput=1

\documentclass[11pt]{article}

\usepackage{naacl2021}

\usepackage{times}
\usepackage{latexsym}

\usepackage[T1]{fontenc}

\usepackage[utf8]{inputenc}

\usepackage{microtype}

\usepackage{graphicx}
\usepackage{amsmath}
\usepackage{amsfonts}
\usepackage{algorithm}
\usepackage{algorithmic}
\usepackage{multirow}
\usepackage{subcaption}
\usepackage{bm}
\usepackage{xcolor}
\usepackage{latexsym}
\usepackage{tabularx,ragged2e}
\newcolumntype{C}{>{\arraybackslash}X}

\newcommand{\beq}{\begin{equation}}
\newcommand{\eeq}{\end{equation}}
\newcommand{\beqs}{\begin{eqnarray}}
\newcommand{\eeqs}{\end{eqnarray}}
\newcommand{\bx}{\bm{x}}

\newcommand{\CL}{\mathcal{L}}

\newcommand{\KL}{\text{KL}}
\newcommand{\EE}{\mathbb{E}}
\newcommand{\BD}{\mathbb{D}}
\newcommand{\BB}{\mathbb{B}}
\newcommand{\Imat}{\boldsymbol{I}}

\newcommand{\CT}{\mathcal{T}}
\newcommand{\CN}{\mathcal{N}}
\newcommand{\muv}{\bm{\mu}}
\newcommand{\phiv}{\bm{\phi}}
\newcommand{\xiv}{\bm{\xi}}

\newcommand{\kappav}{\bm{\kappa}}

\newcommand{\Sigmav}{\bm{\Sigma}}

\newcommand{\CM}{\mathcal{M}}

\newcommand{\uv}{\bm{u}}
\newcommand{\vv}{\bm{v}}
\newcommand{\xv}{\bm{x}}
\newcommand{\yv}{\bm{y}}
\newcommand{\zv}{\bm{z}}
\newcommand{\BR}{\mathbb{R}}
\newcommand{\Gmat}{\bm{G}}

\newcommand{\zerov}{\bm{0}}

\newcommand{\ud}{\,\text{d}}

\newcommand{\argmin}{\arg\min}

\newcommand{\model}{APo-VAE}
\title{APo-VAE: Text Generation in Hyperbolic Space}

\author{Shuyang Dai$^1$\thanks{$\,\,\,$Work was done when the author interned at Microsoft.}
\hspace{0.03in} Zhe Gan$^2$
\hspace{0.03in} Yu Cheng$^2$
\hspace{0.03in} Chenyang Tao$^1$
\hspace{0.03in} Lawrence Carin$^1$
\hspace{0.03in} Jingjing Liu$^2$
\vspace{1mm} \\
\hspace{0.15in} $^{1}$Duke University
\hspace{0.15in} $^{2}$Microsoft  Corporation
\vspace{1mm} \\
\{shuyang.dai, chenyang.tao, lcarin\}@duke.edu\\
\{zhe.gan, yu.cheng, jingjl\}@microsoft.com
}

\begin{document}
\maketitle
\begin{abstract}
Natural language often exhibits inherent hierarchical structure ingrained with complex syntax and semantics. However, most state-of-the-art deep generative models learn embeddings only in Euclidean vector space, without accounting for this structural property of language. We investigate text generation in a hyperbolic latent space to learn continuous hierarchical representations. An Adversarial Poincar{\'e} Variational Autoencoder (APo-VAE) is presented, where both the prior and variational posterior of latent variables are defined over a Poincar{\'e} ball via wrapped normal distributions. By adopting the primal-dual formulation of Kullback-Leibler divergence, an adversarial learning procedure is introduced to empower robust model training. Extensive experiments in language modeling, unaligned style transfer, and dialog-response generation demonstrate the effectiveness of the proposed APo-VAE model over VAEs in Euclidean latent space, thanks to its superb capabilities in capturing latent language hierarchies in hyperbolic space.
\end{abstract}

\section{Introduction}
\label{sec:Intro}
%


\begin{figure}[!t]
    \centering
    \includegraphics[width=0.4\textwidth]{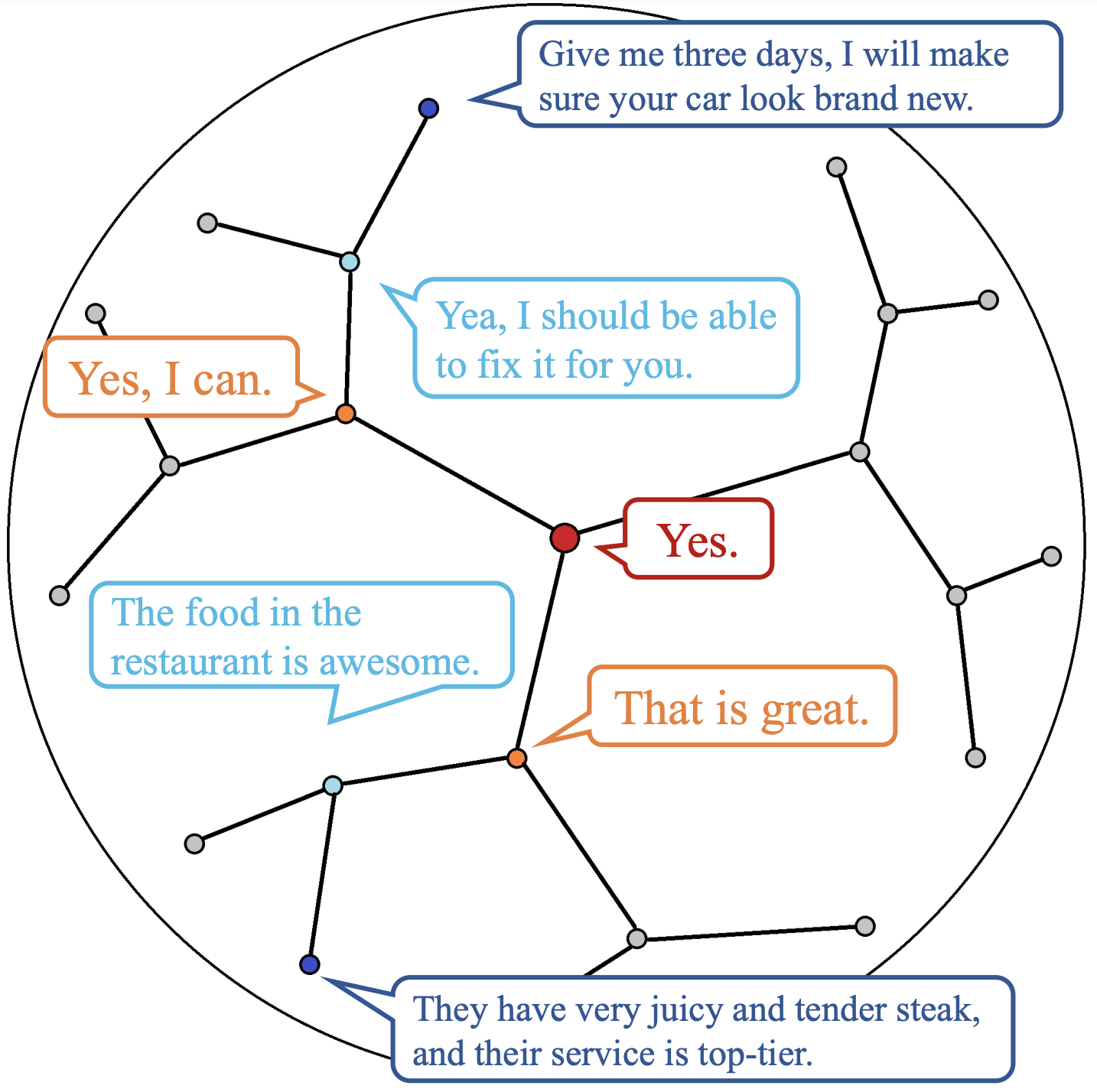}
    \caption{Illustration of the latent hierarchy in natural language. Each tree node is a latent code of its corresponding sentence.}
    \label{fig:demo}
\end{figure}

The Variational Autoencoder (VAE)~\cite{kingma2013auto,rezende2014stochastic} 
is a generative model widely applied to language-generation tasks, which propagates latent codes drawn from a simple prior to manifest data samples through a decoder. The generative model is augmented by an inference network, which feeds observed data samples through an encoder to yield
a distribution on the corresponding latent codes. Since natural language often manifests a latent hierarchical structure, it is desirable for the latent code in a VAE to reflect such inherent language structure, so that the generated text can be more natural and expressive.  
An example of language structure is illustrated in Figure~\ref{fig:demo}, where sentences are organized into a tree structure. The root node corresponds to simple sentences (\emph{e.g.}, ``\emph{Yes}''), while nodes on outer leaves represent sentences with more complex syntactic structure and richer, more specific semantic meaning (\emph{e.g.}, ``\emph{The food in the restaurant is awesome}'')\footnote{Another possible way to organize sentences is a hierarchy of topics, \emph{e.g.}, a parent node can be a sentence on ``\emph{sports}'', while its children are sentences on ``\emph{basketball}'' and ``\emph{skiing}''.}.

In existing VAE-based generative models, such structures are not {\it explicitly} considered. The latent code often employs a simple Gaussian prior, and the posterior is approximated as a Gaussian with diagonal covariance matrix. Such embeddings assume Euclidean structure, which is inadequate in capturing geometric structure illustrated in Figure~\ref{fig:demo}. While some variants have been proposed to enrich the prior distributions~\cite{xu2018spherical,wang2019topic,wang2019improving,shi2019fixing}, there is no evidence that structural information in language can be recovered effectively by the model. 

Hyperbolic geometry has recently emerged as an effective method for representation learning from data with hierarchical structure~\cite{mathieu2019hierarchical,nickel2017poincare}. 
Informally, hyperbolic space can be considered as a continuous map of trees. For example, a Poincar{\'e} disk (a hyperbolic space with two dimensions) can represent any tree with arbitrary low distortion~\cite{de2018representation, sarkar2011low}. In Euclidean space, however, it is difficult to learn such structural representation even with infinite dimensions~\cite{linial1995geometry}.

Motivated by these observations, we propose Adversarial Poincar{\'e} Variational Autoencoder (APo-VAE), a text embedding and generation model based on hyperbolic representations, 
where the latent code is encouraged to capture the underlying tree-like structure in language.
Such latent structure provides more control of the generated sentences, $i.e.$, an increase of sentence complexity and diversity can be achieved along some trajectory from a root to its children.
In practice, we define both the prior and the variational posterior of the latent code over a Poincar{\'e} ball, via the use of a wrapped normal distribution~\cite{nagano2019wrapped}. 
To obtain more stable model training and learn more flexible representation of the latent code, we exploit the primal-dual formulation of Kullback-Leibler (KL) divergence~\cite{dai2018coupled} based on the Fenchel duality~\cite{rockafellar1966extension}, to adversarially optimize the variational bound. Unlike the primal form that relies on Monte Carlo approximation~\cite{mathieu2019hierarchical}, our dual formulation bypasses the need for tractable posterior likelihoods via the introduction of an auxiliary dual function. 

We apply the proposed approach to language modeling, unaligned style transfer and dialog-response generation. For language modeling, in order to enhance the distribution complexity of the prior, we use an additional ``variational mixture of posteriors'' prior (VampPrior) design~\cite{tomczak2017vae} for the wrapped normal distribution. Specifically, VampPrior uses a mixture distribution with components from variational posteriors, coupling the parameters of the prior and variational posterior. 
For unaligned style transfer, we add a sentiment classifier to our model, and disentangle content and sentiment information by using adversarial training~\cite{zhao2017adversarially}.
For dialog-response generation, a conditional model variant of APo-VAE is designed to take into account the dialog context. 


Experiments also show that the proposed model addresses {\it posterior collapse}~\cite{bowman2015generating},
a major obstacle preventing efficient learning of a VAE on text data. In posterior collapse, the encoder learns an approximate posterior similar to the prior, and the decoder tends to ignore the latent code for generation. Experiments show that our proposed model can effectively avoid posterior collapse. We hypothesize that this is due to the use of a more informative prior in hyperbolic space that enhances the complexity of the latent representation, which aligns well with previous work~\cite{tomczak2017vae,wang2019topic} that advocates a better prior design.

Our main contributions are summarized as follows. ($i$) We present Adversarial Poincar{\'e} Variational Autoencoder (APo-VAE), a novel approach to text embedding and generation based on hyperbolic latent representations. ($ii$) In addition to the use of a wrapped normal distribution, an adversarial learning procedure and a VampPrior design are incorporated for robust model training. ($iii$) Experiments on language modeling, unaligned style transfer, and dialog-response generation benchmarks demonstrate the superiority of the proposed approach compared to Euclidean VAEs, as it benefits from capturing informative latent hierarchies in natural language. 



\section{Preliminaries}
\label{sec:Prelim}
\subsection{Variational Autoencoder}
Let $\bm{X}=\{\bm{x}_i\}_{i=1}^N$ be a dataset of sentences, where each  $\bm{x}_i=[x_{i,1},...,x_{i,T_i}]$ is a sequence of tokens of length $T_i$. Our goal is to learn $p_{\bm \theta}(\bx)$ that best models the observed sentences so that the expected log-likelihood is maximized, \textit{i.e.},
$\CL(\bm \theta) = \frac{1}{N} \sum_i \log p_{\bm \theta}(\bx_i)$.

The variational autoencoder (VAE)~\cite{kingma2013auto, chen2018variational} considers a latent-variable model $p_{\bm \theta}(\xv,\zv)$ to represent sentences, with an auxilary encoder 
that draws samples of latent code $\zv$ from the conditional density $q_{\bm \phi}(\zv|\xv)$, known as the approximate posterior. 
Given a latent code $\zv$, the decoder samples a sentence from the conditional density
$p_{\bm \theta}(\xv|\zv) = \prod_t p(\xv_{t} | \xv_{<t}, \zv)$, 
where the ``decoding'' pass takes an auto-regressive form. Together with prior $p(\zv)$, the model is given by the joint $p_{\bm \theta}(\xv,\zv) = p_{\bm \theta}(\xv|\zv)p(\zv)$. 
The VAE leverages the approximate posterior to derive an {\it evidence lower bound} (ELBO) to the (intractable) marginal log-likelihood $\log p_{\bm \theta}(\xv) = \log \int p_{\bm \theta}(\xv,\zv) \ud \zv$:
\beq\label{eq:marginal}
\CL(\xv;\bm \theta,\bm \phi) =
\EE_{\zv\sim q_{\bm \phi}(\zv|\xv)} \left[ \log \frac{p_{\bm \theta}(\xv, \zv)}{q_{\bm \phi}(\zv|\xv)}\right]\,,
\eeq
where $(\bm \theta, \bm \phi)$ are jointly optimized during training, and the gap is given by the decomposition
\beq\label{eq:elbo}
\log p_{\bm \theta}(\xv) = \CL(\xv;\bm \theta,\bm \phi) + \BD_{\KL}(p_{\bm \theta}(\zv|\xv)\parallel q_{\bm \phi}(\zv|\xv)) \,,
\eeq
where $\BD_{\KL}$ denotes Kullback-Leibler divergence. Alternatively, the ELBO can be written as:
\begin{align}\label{eq:vae_obj}
\CL(\xv;\bm \theta,\bm \phi) = &~\EE_{\zv\sim q_{\bm \phi}(\zv|\xv)}\left[ \log p_{\bm \theta}(\xv|\zv) \right] \nonumber\\ &-\BD_{\KL}(q_{\bm \phi}(\zv|\xv)\parallel p(\zv))\,,
\end{align}
where the first conditional likelihood and second KL terms respectively characterize reconstruction and generalization capabilities. 
Intuitively, a good model is expected to strike a balance between good reconstruction and generalization. In most cases, both the prior and  variational posterior are assumed to be Gaussian for computational convenience. However, such over-simplified assumptions may not be ideal for capturing the intrinsic characteristics of data that have unique geometrical structure, such as natural language.

\subsection{Hyperbolic Space}


Riemannian manifolds can provide a 
more powerful and meaningful embedding space for complex data with highly non-Euclidean structure, that cannot be effectively captured in a vectorial form (\textit{e.g.}, social networks, biology and computer graphics). Of particular interest is the hyperbolic space \citep{ganea2018hyperbolic}, where ($i$) the relatively simple geometry allows tractable computations, and ($ii$) the exponential growth of distance in finite dimensions naturally embeds rich hierarchical structure in a compact form. 

\begin{figure*}
    \centering
    \includegraphics[width=0.96\textwidth]{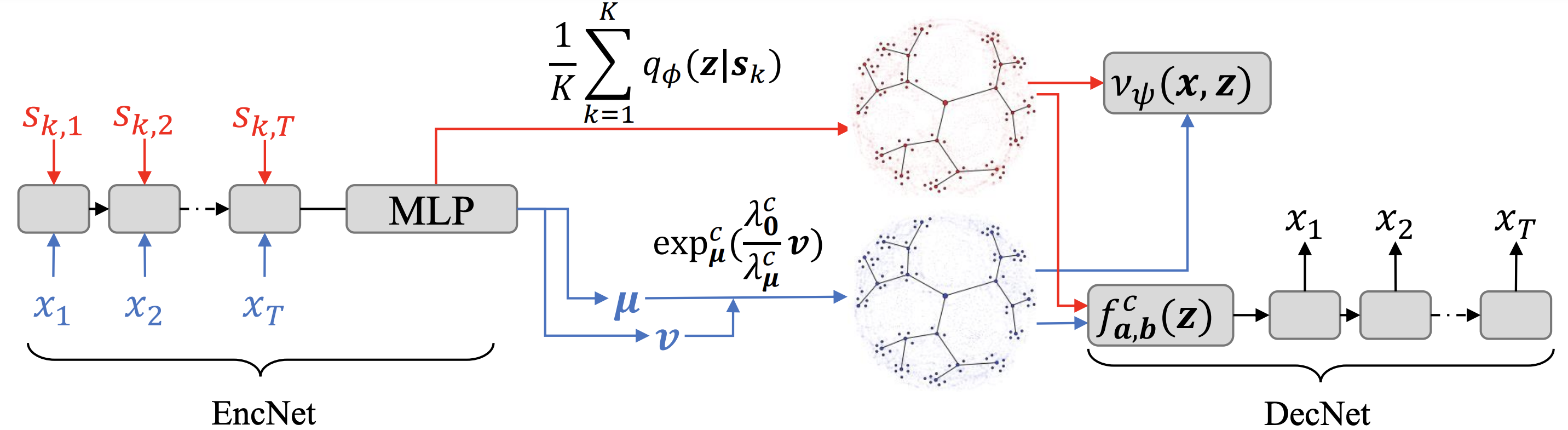}
    \caption{Model framework of the proposed \model~({\color{red}red} is the prior and {\color{blue}blue} is the posterior). $\bm x=[x_1,...,x_T]$ is text sequential data, and $\bm s_k=[s_{k,1},...,s_{k,T}]$ is the pseudo-input. The posterior ({\color{blue}blue}) is obtained by \eqref{eq:reparametrization}, and VampPrior ({\color{red}red}) is achieved by \eqref{eq:vamp_prior}. $\nu_{\bm \psi}(\bm x, \bm z)$ is the dual function.}
    \label{fig:framework}
\end{figure*}

\paragraph{Riemannian Geometry.} An $n$-dimensional Riemannian manifold $\CM^n$ is a set of points locally similar to a linear space $\BR^n$. At each point $\xv$ of the manifold $\CM^n$, we can define a real vector space $\CT_{\xv} \CM^n$ that is tangent to $\xv$, along with an associated {\it metric tensor} $g_{\xv}(\cdot,\cdot):\CT_{\xv}\CM^n\times\CT_{\xv}\CM^n\rightarrow \BR$ which is an inner product on $\CT_{\xv}\CM^n$. Intuitively, a Riemannian manifold behaves like a vector space only in its infinitesimal neighborhood, allowing the generalization of common notation like angle, straight line and distance to a smooth manifold. For each tangent space $\CT_{\xv} \CM^n$, there exists a specific one-to-one map $\exp_{\xv}(\vv):\CT_{\xv}\CM^n \rightarrow \CM^n$ from an $\epsilon$-ball at the origin of $\CT_{\xv} \CM^n$ to a neighborhood of $\xv$ on $\CM^n$, called the {\it exponential map}. 
We refer to the inverse of an exponential map as the {\it logarithm map}, denoted $\log_{\xv}(\yv):\CM^n\rightarrow \CT_{\xv}\CM^n$. In addition, a parallel transport $P_{\bm \xv \rightarrow \bm \xv'}:\CT_{\xv}\CM^n \rightarrow \CT_{\xv'}\CM^n$ intuitively transports tangent vectors along a ``straight'' line between $\xv$ and $\xv'$, so that they remain ``parallel.'' This is the basic machinery that allows us to generalize distributions and computations in the hyperbolic space, as detailed in later sections. 


\paragraph{Poincar{\'e} Ball Model.} Hyperbolic geometry is one type of non-Euclidean geometry with a constant negative curvature. As a classical example of hyperbolic space, an $n$-dimensional Poincar\'e ball, with curvature parameter $c\geq0$ (\textit{i.e.}, radius $\frac{1}{\sqrt{c}}$), can be denoted as $\BB_c^n := \left\{ \zv\in \BR^n \, | \, c \| \zv \|^2 <1  \right\}$ with its metric tensor given by $g_{\zv}^{c} = \lambda_{\zv}^2 g^E$, where $\lambda_{\zv} = \frac{2}{1-c\|\zv\|^2}$ and $g^E$ denotes the regular Euclidean metric tensor. Intuitively, as $\zv$ moves closer to the boundary $\frac{1}{\sqrt{c}}$, the hyperbolic distance between $\zv$ and a nearby $\zv'$ diverges at a rate of $\frac{1}{1-c\|\zv\|^2} \rightarrow \infty$. This implies significant representation capacity, as very dissimilar objects can be encoded on a compact domain. 
Note that as $c\rightarrow 0$, the model recovers the Euclidean space $\BR^n$, $i.e.$, the lack of hierarchy. In comparison, a larger $c$ implies a stronger hierarchical organization.\footnote{The fact that \model~outperforms standard VAE evidences the existence of the hierarchical organization in NLP data.}


%

\paragraph{Mathematical Operations.} We review the closed-form mathematical operations that enable differentiable training for hyperbolic space models, namely the hyperbolic algebra (vector addition) and tangent space computations (exponential/logarithm map and parallel transport). 
The hyperbolic algebra is formulated under the framework of {\it gyrovector spaces} \citep{ungar2008gyrovector}, with the addition of two points $\zv, \zv' \in \BB_c^n$ given by the {\it M\"obius addition}:
\begin{align}\label{eq:mobius_add}
    &\bm z \oplus_c \bm z' := \\
    & \hspace{10pt} \frac{(1+2c\langle\bm z, \bm z'\rangle+c\Vert\bm z'\Vert^2)\bm z+(1-c\Vert\bm z\Vert^2)\bm z'}{1+2c\langle\bm z, \bm z'\rangle+c^2\Vert\bm z\Vert^2\Vert\bm z'\Vert^2}.\nonumber
\end{align}
For any point $\bm \mu\in\BB_c^n$, the exponential map and the logarithmic map are given for $\uv \neq \bm 0$ and $\bm y \neq \bm \mu$ by
\beq
\exp_{\bm \mu}^c(\bm u):=\bm \mu \oplus_c(\tanh(\sqrt{c}\frac{\lambda_{\bm \mu}^c\Vert \bm u \Vert}{2})\frac{\bm u}{\sqrt{c}\Vert \bm u \Vert}),
\nonumber
\eeq
\vspace{-7mm}
\beq
\label{eq:exp_map}
\log_{\bm \mu}^c(\bm y)
    := \frac{2}{\sqrt{c}\lambda_{\bm \mu}^c}\tanh^{-1}(\sqrt{c}\Vert\kappav_{\muv,\yv} \Vert)\frac{\kappav_{\muv,\yv} }{\Vert \kappav_{\muv,\yv} \Vert}, \\
\eeq
where $\kappav_{\muv,\yv}:=(-\bm \mu) \oplus_c \bm y$. Note that the Poincar{\'e} ball model is {\it geodesically complete} in the sense that $\exp^c_{\muv}$ is well-defined on the full tangent space $\CT_{\muv}\BB^n_c$. 
The parallel transport map from a vector $\vv\in\CT_{\zerov} \BB^n_c$ to another tangent space $\CT_{\muv}\BB_c^n$ is given by
\begin{equation}\label{eq:parallel_transport}
    P_{\bm 0 \rightarrow \bm \mu}^c(\bm v)=\log_{\bm \mu}^c(\bm \mu \oplus_c \exp_{\bm 0}^c(\bm v)) = \frac{\lambda_{\bm 0}^c}{\lambda_{\bm \mu}^c}\bm v .
\end{equation}

\section{Adversarial Poincar{\'e} VAE}
\label{sec:Method}
We first introduce our hyperbolic encoder and decoder, and how to apply reparametrization. We then provide detailed descriptions on model implementation, explaining how the primal-dual form of KL divergence can help stabilize training. Finally, we describe how to adopt VampPrior~\cite{tomczak2017vae} to enhance performance. A summary of our model scheme is provided in Figure~\ref{fig:framework}.

\subsection{Flexible Wrapped Distribution Encoder}
We begin by generalizing the standard normal distribution to a Poincar{\'e} ball \citep{ganea2018hyperbolic}. While there are a few competing definitions of the hyperbolic normal, we choose the wrapped normal as our prior and variational posterior, largely due to its flexibility for more expressive generalization. 
A wrapped normal distribution $\CN_{\BB_c^n}(\muv, \Sigmav)$ is defined as follows: ($i$) sample vector $\vv$ from $\CN(\bm 0,\Sigmav)$, ($ii$) parallel transport $\vv$ to $\uv:=P_{\zerov\rightarrow \muv}^c(\vv)$, and ($iii$) using exponential map to project $\uv$ back to $\zv:=\exp_{\muv}^c(\uv)$. Putting these together, a latent sample has the following reparametrizable form: 
\beq \label{eq:reparametrization}
\zv = \exp_{\muv}^c\left( \frac{\lambda_{\zerov}^c}{\lambda_{\muv}^c} \vv \right), \vv \sim \CN(\zerov,\Sigmav).
\eeq
For approximate posteriors, $(\muv,\Sigmav)$ depends on $\xv$. 
%
We further generalize the (restrictive) hyperbolic wrapped normal by acknowledging that under the implicit VAE~\cite{fang2019implicit} framework, one does not need the approximate posterior $q_{\bm \phi}(\zv|\xv)$ to be analytically tractable. This allows us to replace the tangent space sampling step $\vv \sim \CN(\zerov,\Sigmav)$ in \eqref{eq:reparametrization} with a more flexible implicit distribution from which we draw samples as $\vv:=G(\xv, \xiv;\phiv_1)$ for $\xiv \sim \CN(\zerov, \Imat)$. Note that now $\muv:=F(\xv;\phiv_2)$ can be regarded as a deterministic displacement vector that anchors embeddings to the correct semantic neighborhood, allowing the stochastic $\vv$ to only focus on modeling the local uncertainty of the semantic embedding. The synergy between the deterministic and stochastic parts enables efficient representation learning relative to existing alternatives. For simplicity, we denote the encoder neural network as $\text{EncNet}_{\bm\phi}$, which contains $G$ and $F$, with parameters $\bm \phi=\{\phiv_1, \phiv_2\}$. 

\subsection{Poincar\'e Decoder}
To build a geometry-aware decoder for a hyperbolic latent code, we follow~\citet{ganea2018hyperbolic}, and use a generalized linear function analogously defined in the hyperbolic space. A Euclidean linear function takes the form $f(\zv) = \langle \bm a, \zv - \bm b\rangle =\text{sign}(\langle \bm a, \zv - \bm b\rangle)\|\bm a \| d^E(\zv, H_{\bm a, \bm b})$, where $\bm a$ is the coefficient, $\bm b$ is the intercept, $H_{\bm a, \bm b}$ is a hyperplane passing through $\bm b$ with $\bm a$ as the normal direction, and $d^E(\zv, H)$ is the distance between $\zv$ and hyperplane $H$. The counterpart in Poincar{\'e} ball analogously writes
\begin{equation}\label{eq:decode}
    f_{\bm a,\bm b}^{c}(\bm z) = \text{sign}(\langle\bm a, \log_{\bm b}^c(\bm z)\rangle_{\bm b})\Vert\bm a\Vert_{\bm b}\,d_c^{\BB}(\bm z, H_{\bm a, \bm b}^c),
\end{equation}
where 
$H_{\bm a, \bm b}^c=\{\bm z\in\BB_c^n|\langle\bm a, \log_{\bm b}^c(\bm z)\rangle_{\bm b}=0\}$, and $d_c^{\BB}(\bm z, H_{\bm a, \bm b}^c)=\frac{1}{\sqrt{c}}\sinh^{-1}\left(\frac{2\sqrt{c}|\langle \kappav_{\bm b,\bm z}, \bm a\rangle|}{(1-c\Vert\kappav_{\bm b,\bm z} \Vert^2)\Vert \bm a\Vert}\right)$
are the the gyroplane and the distance between $\bm z$ and the gyroplane, respectively.  
Specifically, we use the hyperbolic linear function in \eqref{eq:decode} to extract features from the Poincar\'e embedding $\zv$. The feature $f_{\bm a,\bm b}^{c}(\bm z)$ will be the input to the RNN decoder. We denote the combined network of $f_{\bm a,\bm b}^{c}$ and the RNN decoder as $\text{DecNet}_{\bm \theta}$, where parameters $\bm \theta$ contain $\bm a$ and $\bm b$.


\subsection{Implementing \model}
While it is straightforward to compute the ELBO \eqref{eq:vae_obj} via Monte Carlo estimates using the explicit wrapped normal density~\cite{mathieu2019hierarchical}, we empirically observe that: ($i$) the normal assumption restricts the expressiveness of the model, and ($ii$) the wrapped normal likelihood makes the training unstable. 
Therefore, we appeal to a primal-dual view of VAE training to overcome such difficulties~\cite{rockafellar1966extension, dai2018coupled, tao2019fenchel, fang2019implicit}. Specifically, the KL term in~\eqref{eq:vae_obj} can be reformulated as:
\begin{align}\label{eq:dual_KL_function}
    &\BD_{\text{KL}}(q_{\bm \phi}(\zv|\xv)\parallel p(\zv))= \max_{\bm\psi} \\
    & \left\{\mathbb{E}_{\bm z\sim q_{\bm \phi}(\bm z| \bm x)}\nu_{\bm \psi}(\bm x, \bm z)-\mathbb{E}_{\bm z\sim p(\bm z)}\exp\nu_{\bm \psi}(\bm x, \bm z) \right\},\nonumber
\end{align}
where $\nu_{\bm \psi}(\bm x, \bm z)$ is the (auxiliary) dual function (\textit{i.e.}, a neural network) with parameters $\bm \psi$. The primal-dual view of the KL term enhances the approximation ability, while also being tractable computationally. Meanwhile, since the density function in the original KL term in~\eqref{eq:vae_obj} is replaced by the dual function $\nu_{\bm \psi}(\bm x, \bm z)$, we can avoid direct computation with respect to the probability density function of the wrapped normal distribution.

To train our proposed \model~with the primal-dual form of the VAE objective, we follow the training schemes of coupled variational Bayes (CVB)~\cite{dai2018coupled} and implicit VAE~\cite{fang2019implicit}, which optimize the objective adversarially. Specifically, we update $\bm \psi$ in the dual function $\nu_{\bm \psi}(\bm x, \bm z)$ to maximize:
\begin{align}\label{eq:dual_KL}
    \mathcal{L}_1 = \mathbb{E}_{\bm x \sim \bm X} [&\,\mathbb{E}_{\bm z\sim q_{\bm \phi}(\bm z| \bm x)}\nu_{\bm \psi}(\bm x, \bm z)\nonumber\\
    &-\mathbb{E}_{\bm z\sim p(\bm z)}\exp\nu_{\bm \psi}(\bm x, \bm z)\,]\,,
\end{align}
where $\mathbb{E}_{\bm x \sim \bm X}[\cdot]$ denotes the expectation over empirical distribution on observations.
Accordingly, parameters $\bm \theta$ and $\bm \phi$ are updated to maximize:
\begin{align}\label{eq:adversarial}
    \mathcal{L}_2 = \mathbb{E}_{\bm x \sim \bm X}\mathbb{E}_{\bm z \sim q_{\bm \phi}(\bm z | \bm x)}[&\,\log p_{\bm \theta}(\bm x| \bm z) \nonumber\\
    &- \nu_{\bm \psi}(\bm x, \bm z)\,].
\end{align}
Note that the term $\mathbb{E}_{\bm x \sim \bm X}\mathbb{E}_{\bm z \sim q_{\bm \phi}(\bm z | \bm x)} \nu_{\bm \psi}(\bm x, \bm z)$ is maximized in~\eqref{eq:dual_KL} while it is minimized in~\eqref{eq:adversarial}, \textit{i.e.}, adversarial learning.
In other words, one can consider the dual function as a discriminative network that distinguishes between the prior $\bm z \sim p(\bm z)$ and the variational posterior $\bm z \sim q_{\bm \phi}(\bm z|\bm x)$, both of which are paired with the input data $\bm x \sim \bm X$.

\begin{algorithm}[t]
\begin{algorithmic}[1]
\STATE \textbf{Input}: Data samples $\bm X = \{\bm x_i\}_{i=1}^N$, Poincar{\'e} curvature $c$, and number of pseudo-input $K$.
\STATE Initialize $\bm \theta$, $\bm \phi$, $\bm \psi$, and $\bm \delta$.
\FOR{$\text{$iter$ from 1 to $max\_iter$}$}
\STATE Sample a mini-batch $\{\bm x_m\}_{m=1}^M$ from $\bm X$ of size $M$.
\vspace{1mm}
\STATE \textit{\textbf{\footnotesize \# Sampling in the Hyperbolic Space.}}
\STATE Obtain $\bm \mu_m$ and $\vv_m$ from $\text{EncNet}_{\bm \phi}(\bm x_m)$.\\
\STATE Move $\bm v_m$ to $\bm u_m=P_{\bm 0\rightarrow \bm \mu_m}^c(\bm v_m)$ by~\eqref{eq:parallel_transport}. 
\STATE Map $\bm u_m$ to $\bm z_m=\exp_{\bm \mu_m}^c(\bm u_m)$ by~\eqref{eq:exp_map}. 
\vspace{1mm}
\STATE \textit{\textbf{\footnotesize \# Update the dual function and the pseudo-input.}}
\STATE Sample $\tilde{\bm z}_m$ by~\eqref{eq:vamp_prior}.\\
\STATE Update $\bm \psi$ and $\bm \delta$ by gradient ascent on~\eqref{eq:dual_KL}
\vspace{1mm}
\STATE \textit{\textbf{\footnotesize \# Update the encoder and decoder networks.}}
\STATE Update $\bm \theta$ and $\bm \phi$ by gradient ascent on~\eqref{eq:adversarial}.
\ENDFOR
\caption{Training procedure of \model.}
\label{alg:main}
\end{algorithmic}
\end{algorithm}

\subsection{Data-driven Prior}
While the use of a standard normal prior is a simple choice in Euclidean space, we argue that it induces bias in the hyperbolic setup. This is because natural sentences have specific meaning, and it is unrealistic to have the bulk of mass concentrated in the center (this is for low dimension; for high dimensions, it will concentrate near the surface of a sphere, which may partly explain why cosine similarity works favorably compared with Euclidean distance for NLP applications). 

To reduce the induced bias from a pre-fixed prior, 
we adopt the VampPrior framework~\cite{tomczak2017vae}, which is a mixture of variational posteriors conditioned on learnable pseudo-data points.
Specifically, we consider the prior as a learnable distribution given by
\begin{equation}\label{eq:vamp_prior}
    p_{\bm \delta}(\bm z) = \frac{1}{K}\textstyle{\sum^{K}_{k=1}}q_{\bm \phi}(\bm z| \bm s_k) , 
\end{equation}
where $q_{\bm \phi}$ is the learned approximate posterior, and we call the parameter $\bm \delta := \{\bm s_k\}_{k=1}^K$ pseudo inputs. Intuitively, $p_{\bm \delta}(\bm z)$ seeks to match the aggregated posterior~\cite{makhzani2015adversarial}: $q(\bm z) = \frac{1}{N}\sum^N_{i=1}q_{\bm \phi}(\bm z| \bm x_i)$ in a cost-efficient manner via parameterizing the pseudo inputs. 
By replacing the prior distribution $p(\zv)$ in~\eqref{eq:dual_KL} with $p_{\bm \delta}(\bm z)$, we complete the final objective of the proposed \model. The detailed training procedure is summarized in Algorithm~\ref{alg:main}.

\section{Related Work}
\label{sec:Related}





\paragraph{VAE for Text Generation.}
Many VAE models have been proposed for text generation, most of which focus on solving the issue of posterior collapse. The most popular strategy is to alter the training dynamics, keeping the encoder away from bad local optima. For example, variants of KL annealing~\cite{bowman2015generating,zhao2017infovae, fu2019cyclical} dynamically adjust the weight on the KL penalty term as training progresses. Lagging VAE~\cite{he2019lagging} aggressively optimizes the encoder before each decoder update, to overcome the imbalanced training issue between the encoder and decoder.  
Alternative strategies have also been proposed based on competing theories or heuristics.  $\delta$-VAE~\cite{razavi2019preventing} tackles this issue by enforcing a minimum KL divergence between the posterior and the prior.  \citet{yang2017improved} blames mode-collapse on the auto-regressive design of the decoder and advocates alternative architectures. 
A semi-amortized inference network is considered by~\citet{kim2018semi} to bridge the amortization gap between log-likelihood and the ELBO.




Recent work has also shown that posterior collapse can be ameliorated by using more expressive priors and variational posteriors other than Gaussian. 
Flow-based VAE is considered in~\citet{ziegler2019latent} to enhance the flexibility of prior distributions. A topic-guided prior is proposed in~\citet{wang2019topic} to achieve more controllable text generation. 
\citet{fang2019implicit} explores implicit sample-based representations, without requiring an explicit density form for the approximate posterior. \citet{xu2018spherical} considers replacing the Gaussian with the spherical von Mises-Fisher (vMF) distribution. Compared to these prior arts, our model features structured representation in hyperbolic space, which not only captures latent hierarchies but also combats posterior collapse.

\paragraph{Hyperbolic Space Representation Learning.} There has been a recent surge of interest in representation learning in hyperbolic space, largely due to its exceptional effectiveness modeling data with underlying graphical structure \citep{chamberlain2017neural}, such as relation nets \citep{nickel2017poincare}. In the context of NLP, hyperbolic geometry has been considered for word embeddings \citep{tifrea2018poincar}. A popular vehicle for hyperbolic representation learning is the auto-encoder (AE) framework \citep{grattarola2019adversarial, ovinnikov2019poincare}, where the decoders are built to efficiently exploit the hyperbolic geometry \citep{ganea2018hyperbolic}. Closest to our APo-VAE are the works of hyperbolic VAEs \citep{mathieu2019hierarchical, nagano2019wrapped}, where wrapped normal distributions have been used. Drawing power from the dual form of the KL, the proposed APo-VAE highlights an implicit posterior and data-driven prior, showing improved training stability.

\section{Experiments} \label{sec:Exp}
We evaluate the proposed model on three tasks: ($i$) language modeling, ($ii$) unaligned style transfer, and ($iii$) dialog-response generation, with quantitative results, human evaluation and qualitative analysis.  

\subsection{Experimental Setup}
\paragraph{Datasets.} 
We use three datasets for language modeling: {\it Penn Treebank} (PTB)~\cite{marcus1993building}, {\it Yahoo} and {\it Yelp} corpora~\cite{yang2017improved}. PTB contains one million words of 1989 Wall Street Journal material annotated in Treebank II style, with 42k sentences of varying lengths. Yahoo and Yelp are much larger datasets, each containing 100k sentences with greater average length.

For unaligned style transfer, we use the Yelp restaurant reviews dataset~\cite{shen2017style}, which is obtained by pre-processing the Yelp dataset, $i.e.$, sentences are shortened for more feasible sentence level sentiment analysis.
Overall, the dataset includes 350k positive and 250k negative reviews (based on user rating). 

Following~\citet{gu2018dialogwae}, we use the Switchboard~\cite{godfrey1997switchboard} dataset for dialogue-response generation.
The former contains 2.4k two-sided telephone conversations, manually transcribed and aligned. 
We split the data into training, validation and test sets following the protocol described in \citet{zhao2017learning}. 

\paragraph{Evaluation Metrics.}
To benchmark language modeling performance, we report the ELBO and {\it Perplexity} (PPL) of \model~and baselines. In order to verify our proposed Apo-VAE is more resistant to posterior collapse, we also report the KL-divergence $\BD_{\KL}(q_{\bm \phi}(\bm z| \bm x)\Vert p(\bm z))$ and {\it mutual information} (MI) between $\bm z$ and $\bm x$~\cite{he2019lagging}. The number of active units (AU) of the latent code is also reported, where the activity of a latent dimension $z$ is measured as $A_{\zv}=\text{Cov}_{\xv}\EE_{\zv\sim q_{\phiv}(\zv|\xv)}[z]$, and defined as active if $A_{\zv} > 0.01$.

To evaluate our model on unaligned style transfer, we consider the transfer accuracy from one sentiment to another, the BLEU scores between original and transferred sentences, the reconstruction perplexity of original sentences, and the reverse perplexity (RPPL) based on a language model from the transferred sentences.

For dialogue-response generation, we adopt the evaluation metrics used in previous studies~\cite{zhao2017learning,gu2018dialogwae}, including BLEU~\cite{papineni-etal-2002-bleu}, BOW~\cite{liu2016not}, and \textit{intra/inter-dist} values~\cite{gu2018dialogwae}. The first two metrics are used to assess the relevance of the generated response, and the third is for diversity evaluation. 

\setlength{\tabcolsep}{6pt}
\begin{table}[t]
\small
  \centering
  \begin{tabular}{l|ccccc}
  \hline
    \multirow{2}{*}{Model} & -ELBO & PPL & KL & MI & AU\\
    \cline{2-6}
    &\multicolumn{4}{c}{PTB}\\
    \hline
    VAE & 102.6 & 108.26 & 1.1 & 0.8 & 2\\
    $\beta$-VAE & 104.5 & 117.92 & 7.5 & 3.1 & 5\\
    SA-VAE & 102.6 & 107.71 & 1.2 & 0.7 & 2\\
    vMF-VAE & 95.8 & 93.70 & 2.9 & 3.2 & 21\\
    $\mathcal{P}$-VAE & 91.4 & 76.13 & 4.5 & 2.9 & 23 \\
    iVAE & 87.2 & 53.44 & \textbf{12.5} & \textbf{12.2} & \textbf{32}\\
    \model & 87.2 & 53.32 & 8.4 & 4.8 & \textbf{32}\\
    \model+VP & \textbf{87.0} & \textbf{53.02} &  8.9 & 4.5& \textbf{32}\\
    \hline
    &\multicolumn{4}{c}{Yahoo}\\
    \hline
    VAE & 328.6 & 61.21 & 0.0 & 0.0 & 0\\
    $\beta$-VAE & 328.7 & 61.29 & 6.3 & 2.8 & 8\\
    SA-VAE & 327.2 & 60.15 & 5.2 & 2.9 & 10\\
    LAG-VAE & 326.7 & 59.77 & 5.7 & 2.9 & 15\\
    vMF-VAE & 318.5 & 53.92 & 6.3 & 3.7 & 23\\
    $\mathcal{P}$-VAE & 313.4 & 50.57 & 7.2 & 3.3 & 27 \\
    iVAE & 309.1 & 47.93 & \textbf{11.4} & \textbf{10.7} & \textbf{32}\\
    \model & 286.2 & 47.00 & 6.9 & 4.1 & \textbf{32}\\
    \model+VP & \textbf{285.6} & \textbf{46.61} & 8.1 & 4.9 & \textbf{32}\\
    \hline
    &\multicolumn{4}{c}{Yelp}\\
    \hline
    VAE & 357.9 & 40.56 & 0.0 & 0.0 & 0\\
    $\beta$-VAE & 358.2 & 40.69 & 4.2 & 2.0 & 4\\ 
    SA-VAE & 357.8 & 40.51 & 2.8 & 1.7 & 8\\
    LAG-VAE & 355.9 & 39.73 & 3.8 & 2.4 & 11\\
    vMF-VAE & 356.2 & 51.03 & 4.1 & 3.9 & 13\\
    $\mathcal{P}$-VAE & 355.4 & 50.64 & 4.3 & 4.8 & 19 \\
    iVAE & 348.7 & 36.88 & 11.6 & \textbf{11.0} & \textbf{32}\\
    \model & 319.7 & 34.10 & 12.1 & 7.5 & \textbf{32}\\
    \model+VP & \textbf{316.4} & \textbf{32.91} & \textbf{12.7} & 6.2 & \textbf{32}\\
  \hline
  \end{tabular}
  \vspace{-2mm}
  \caption{Results on PTB, Yahoo, and Yelp datasets. A better language model achieves lower negative ELBO and PPL. Higher KL and MI indicate a better utilization of the latent space.} 
  \label{tab:language_result}
\end{table}


\paragraph{Model Implementation.}
For language modeling, we adopt the LSTM~\cite{hochreiter1997long} for both the encoder and decoder, with dimension of the latent code set to $32$. Following~\citet{mathieu2019hierarchical}, the hyper-parameter $c$ is set to $0.7$. For unaligned style transfer, we extend our model in the same fashion as~\citet{fang2019implicit}.
For dialogue-response generation, we modify \model~following the conditional VAE framework~\cite{zhao2017learning}. Specifically, an extra input of context embedding $\bm s$ is supplied to the model ({\it i.e.}, $p_{\theta}(\xv,\zv|\bm s), q_{\phi}(\zv|\xv,\bm s)$). The prior $p(\bm z| \bm s)$ is a wrapped normal conditioned on context embedding, learned together with the posterior. 

\begin{figure}[!t]
    \centering
    \includegraphics[width=0.48\textwidth]{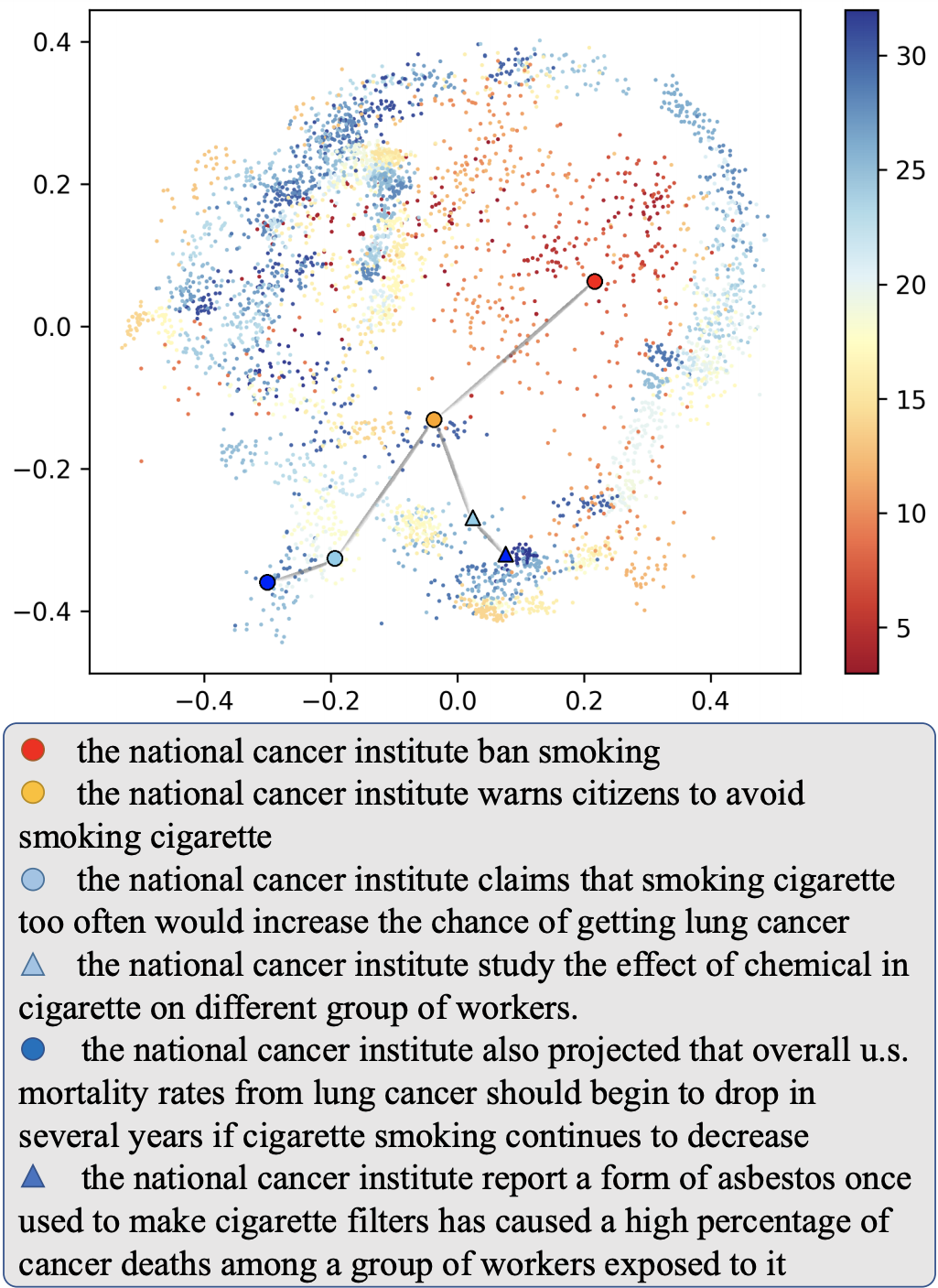}
    \caption{Visualization of the hyperbolic latent space of 5,000 randomly sampled sentences from the test set of PTB. The lengths of the samples are color-coded ({\color{red}red} for short ones and {\color{blue}blue} for longer ones). The six listed sentences are created by modifying data samples.}
    \label{fig:visualization}
\end{figure}

\setlength{\tabcolsep}{9pt}
\begin{table*}
    \small
    \centering
    \begin{tabular}{l|ccc|ccc|cc|cc}
    \hline
        \multirow{2}{*}{Model} & \multicolumn{3}{|c|}{BLEU} & \multicolumn{3}{|c|}{BOW} & \multicolumn{2}{|c|}{Intra-dist} & \multicolumn{2}{|c}{Inter-dist}\\
        \cline{2-11}
        & R & P & F1 & A & E & G & dist-1 & dist-2 & dist-1 & dist-2\\
        \hline
        CVAE & 0.295 & 0.258 & 0.275 & 0.836 & 0.572 & 0.846 & 0.803 & 0.415 & 0.112 & 0.102\\
        CVAE-BOW & 0.298 & \textbf{0.272} & 0.284 & 0.828 & 0.555 & 0.840 & 0.819 & 0.493 & 0.107 & 0.099\\   
        CVAE-CO & 0.299 & 0.269 & 0.283 & 0.839 & 0.557 & 0.855 & 0.863 & 0.581 & 0.111 & 0.110\\
        DialogWAE & 0.394 & 0.254 & 0.309 & 0.897 & 0.627 & 0.887 & 0.713 & 0.651 & 0.245 & 0.413\\
        iVAE & 0.427 & 0.254 & 0.319 & 0.930 & 0.670 & 0.900 & 0.828 & 0.692 & 0.391 & 0.668\\
        \model & \textbf{0.438} & 0.261 & \textbf{0.328} & \textbf{0.937} & \textbf{0.683} & \textbf{0.904} & \textbf{0.861} & \textbf{0.792} & \textbf{0.445} & \textbf{0.717}\\
    \hline
    \end{tabular}
    \vspace{-2mm}
    \caption{Results on SwitchBoard (P: precision, R: recall, A: average, E: extreme, G: greedy). Higher BLEU and BOW Embedding indicate better quality of generated responses. Higher intra/inter-dist means better generation diversity.}
    \label{tab:dialogue_result}
\end{table*}

\subsection{Experimental Results}
\paragraph{Language Modeling.}
Table~\ref{tab:language_result} shows results on language modeling. We compare \model~with other VAE-based solutions, including $\beta$-VAE~\cite{higgins2017beta}, SA-VAE~\cite{kim2018semi}, lagging VAE (LAG-VAE)~\cite{he2019lagging}, vMF-VAE~\cite{xu2018spherical}, Poincar\'e VAE ($\mathcal{P}$-VAE)~\cite{mathieu2019hierarchical} and iVAE\footnote{We report iVAE$_{\text{MI}}$ results in all our experiments.}~\cite{fang2019implicit}. On all three datasets, the proposed model achieves lower negative ELBO and PPL than other models, demonstrating its strong ability to better model sequential text data. Meanwhile, the larger KL term and higher mutual information (between $\bm z$ and $\bm x$) of \model~model indicate its robustness in handling posterior collapse. In addition, the introduction of a data-driven prior (denoted as \model+VP) further boosts the performance, especially on negative ELBO and PPL. 


\paragraph{Visualization.}
To verify our hypothesis that the proposed model is capable of learning latent tree structure in text data, we visualize the two-dimensional projection of the learned latent code in Figure~\ref{fig:visualization}. For visualization, we randomly draw 5k samples from PTB-test, and encode them to the latent space using the \model~encoder. We color-code each sentence based on its length (\textit{i.e.}, blue for long sentences and red for short sentences). 
Note that only a small portion of data have a length longer than $32$ ($<10\%$), and human inspection verified that most of them contain multiple sub-sentences.
We exclude these samples from our analysis. 

\setlength{\tabcolsep}{10pt}
\begin{table}[!t]
\small
    \centering
    \begin{tabular}{l|cccc}
    \hline
        Model & ACC & BLEU & PPL & RPPL \\
        \hline
        ARAE & \textbf{95} & 32.5 & 6.8 & 395 \\
        iVAE & 92 & 36.7 & 6.2 & 285\\
        APo-VAE & \textbf{95} & \textbf{37.8} & \textbf{6.1} & \textbf{273}\\
        \hline
    \end{tabular}
    \vspace{-2mm}
    \caption{Unaligned style transfer on the Yelp restaurant reviews dataset.}
    \label{tab:style_transfer}
\end{table}

\setlength{\tabcolsep}{4.5pt}
\begin{table}[!t]
\small
\centering
\begin{tabular}{l|ccc|ccc}
\hline
\multirow{2}{*}{} & \multicolumn{3}{c|}{vs\, iVAE} & \multicolumn{3}{c}{vs\, DialogWAE} \\
\cline{2-7}
 & win & loss & tie & win & loss & tie \\ 
\hline
Informativeness & 52.8 & 27.9 & 19.3 & 63.7 & 27.1 & 19.2 \\
Coherence & 41.7 & 35.5 & 22.8 & 41.2 & 34.4 & 24.4 \\
Diversity & 51.2 & 26.4 & 22.4 & 62.1 & 25.1 & 12.8 \\
\hline
\end{tabular}
\vspace{-2mm}
\caption{Human evaluation results. Win/loss/tie indicates the percentage of responses generated by APo-VAE being better/worse/equal to the compared model.}
\label{tab:human_eval}
\end{table}

As shown in Figure~\ref{fig:visualization}, longer sentences (dark blue) tend to occupy the outer rim of the Poincar\'e ball, while the shorter ones (dark red) are concentrated in the inner area. 
We also select some long sample sentences (dark blue), and manually shorten them to create several variants of different lengths (ranging from 6 to 27), which are related in a hierarchical manner based on human judgement. We visualize their latent codes projected by the trained \model. The resulting plot is consistent with a hierarchical structure for \model: as the sentence becomes more specific, the embedding moves outward. We also decode from the neighbours of these latent codes, the outputs (see the Appendix) of which demonstrate a similar hierarchical structure.


\paragraph{Unaligned Style Transfer.} Table~\ref{tab:style_transfer} shows the results on the Yelp restaurant reviews dataset. APo-VAE achieves over 1 point increased BLEU scores than iVAE, capturing a more informative and structured feature space. Comparable performance is achieved for the other evaluation metrics.


\paragraph{Dialogue Response Generation.}
Results on SwitchBoard are summarized in Table~\ref{tab:dialogue_result}. Our proposed model generates comparable or better responses than the baseline models in terms of both relevance (BLEU and BOW) and diversity (intra/inter-dist). \model~improves the average recall from $0.427$ (by iVAE) to $0.438$, while significantly enhancing generation diversity (\textit{e.g.}, from $0.692$ to $0.792$ for intra-dist-2). 


\paragraph{Human Evaluation.}
We further perform human evaluation via Amazon Mechanical Turk. We asked the turkers to compare generated responses from two models, and assess each model's informativeness, relevance to the dialog context (coherence), and diversity. We use $500$ randomly sampled contexts from the test set, each assessed by three judges. In order to evaluate diversity, 5 responses are generated for each dialog context. For quality control, only workers with a lifetime task approval rating
greater than 98\% were allowed to participate in our study. Table~\ref{tab:human_eval} summarizes the human evaluation results. The responses generated by our model are clearly preferred by the judges compared with other competing methods.



\section{Conclusions}
\label{sec:Conclusion}
We present \model, a novel model for text generation in hyperbolic space. Our model can learn latent hierarchies in natural language via the use of wrapped normals for the prior. 
A primal-dual view of KL divergence is adopted for robust model training. 
Extensive experiments on language modeling, text style transfer, and dialog response generation demonstrate the superiority of the model. For future work, we plan to combine \model~with the currently prevailing large-scale pre-trained language models. 

\nocite{shao2020controlvae}
\nocite{mescheder2017adversarial}
\nocite{chen2018adversarial}

\bibliography{anthology,custom}
\bibliographystyle{acl_natbib}

\newpage

\renewcommand{\UrlFont}{\ttfamily\small}
\renewcommand\thesection{\Alph{section}}
\renewcommand\thesubsection{\thesection.\arabic{subsection}}
\renewcommand{\thetable}{A\arabic{table}}   
\renewcommand{\thefigure}{A\arabic{figure}}
\renewcommand{\theequation}{A\arabic{equation}}
\setcounter{section}{0}


\section{Basics of Riemannian Geometry (Extended)}

This section provides additional coverage of the basic mathematical concepts used in APo-VAE. For more detailed mathematics expositions, readers are referred to \citet{ganea2018hyperbolic}. 

A real, smooth $n$-dimensional manifold $\CM$ is a set of points locally similar to a linear space $\BR^n$. At each point $\xv$ of the manifold $\CM$ is defined a real vector space of the space of the same dimensionality as $\CM$, called the tangent space in $\xv$: $\CT_{\xv} \CM$. Intuitively it contains all the possible directions in which one can tangentially pass through $\xv$. For each point $\xv$ there also defines a {\it metric tensor} $g_{\xv}(\cdot,\cdot):\CT_{\xv}\CM\times\CT_{\xv}\CM\rightarrow \BR$, which defined an inner product on the associated tangent space $\CT_{\xv}\CM$. More specifically, given a coordinate system, the inner product is given in the quadratic form $g_{\xv}(\uv, \vv)=\langle \uv, \vv \rangle_{g_{\xv}} = \uv^T \Gmat_{\xv} \vv$, where by slight abuse of notation $\uv, \vv \in \BR^n$ are vector representations of the tangent vectors wrt the coordinate system and $\Gmat_{\xv} \in \BR^{n\times n}$ is a positive-definite matrix. Collectively, $(\CM, g)$ defines a {\it Riemannian manifold}. 


\begin{figure}[h]
\centering
\begin{subfigure}{.225\textwidth}
  \centering
  \vspace{-4mm}
  \includegraphics[width=.99\linewidth]{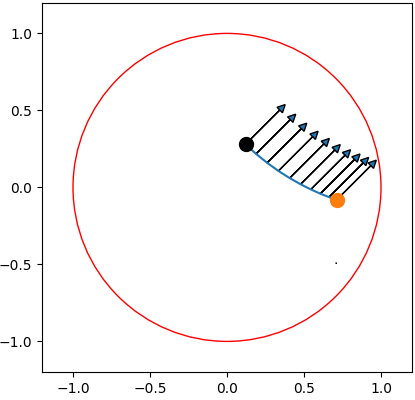}
  \caption{Parallel transport a tangent vector (arrow) from one location (black) to another (orange).}
  \label{fig:sub1}
\end{subfigure}%
\hspace{2mm}
\begin{subfigure}{.225\textwidth}
  \centering
  \includegraphics[width=.99\linewidth]{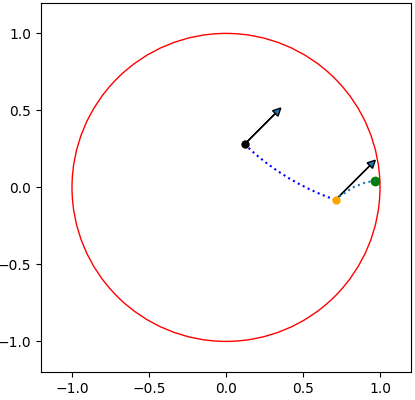}
  \caption{Map the transported tangent vector (orange) to a point (green) in the hyperbolic space by using the exponential map.}
  \label{fig:sub2}
\end{subfigure}
\caption{Visualization of different mathematical operations in a hyperbolic space, that are used to define the wrapped distribution.}
\label{fig:test}
\vspace{-5mm}
\end{figure}

The metric tensor is used to generalize the notations such as distance and volume in Euclidean space to the Riemannian manifold. Given a curve $\gamma(t):[0,1]\rightarrow\CM$, its length is given by
\beq
L(\gamma) =\int_{0}^1 \| \gamma'(t) \|_{\gamma(t)} \ud t\,.
\eeq
The concept of straight lines can then be generalized to {\it geodesics}, which is the shortest path between pairs of points $\xv, \yv$ on the manifold $\gamma^*(\xv,\yv) = \argmin_{\gamma} L(\gamma)$ such that $\gamma(0) = \xv$ and $\gamma(1) = \yv$ with $\gamma$ traveling at constant speed ({\it i.e.}, $\| \gamma'(t) \|_{\gamma(t)}=c$, where $c$ is the distance).  The concept of moving along a straight line with constant speed defines the {\it exponential map}, where for $\vv\in \CT_{\xv}\CM$ there is a unique unit speed geodesic $\gamma$ satisfying $\gamma(0) = \xv$ with initial tangent vector $\vv$, and the corresponding exponential map is defined by $\exp_{\xv}(\vv)=\gamma(1)$. We call the inverse of exponential map the {\it logarithm map} $\log_{\xv} = \exp_{\xv}^{-1}:\CM \rightarrow \CT_{\xv} \CM$, mapping from the manifold to the tangent space. For the Poinc\'are ball model, it is {\it geodesically complete} in the sense that $\exp_{\xv}$ is well-defined on the full tangent space $\CT_{\xv}\CM$. 
\begin{figure}[t]
    \centering
    \includegraphics[width=0.49\textwidth]{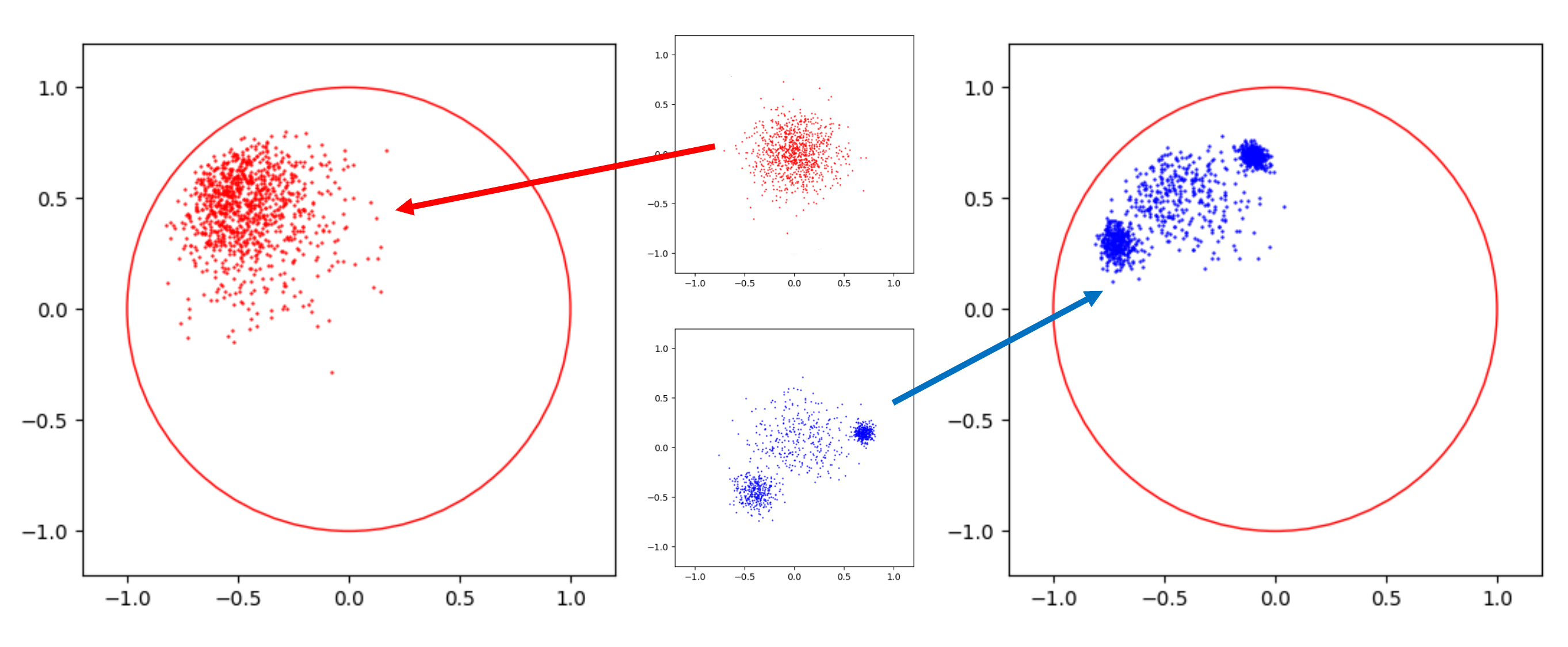}
    \caption{Mapping a Gaussian distribution (red) and a implicit distribution (blue) to the hyperbolic space.}
    \label{fig:map}
    \vspace{-5mm}
\end{figure}

\section{Additional Related Work}

\paragraph{VAE with Adversarial Learning.}
One of the first to apply adversarial learning to VAE is Adversarial Variational Bayes (AVB)~\cite{mescheder2017adversarial,pu2017adversarial}. Motivated by Generative Adversarial Network (GAN)~\cite{goodfellow2014generative}, AVB introduces an auxiliary discriminator that transforms the maximum-likelihood-problem into a two-player game. Similarly, Adversarial Autoencoder (AAE)~\cite{makhzani2015adversarial} uses adversarial learning to match aggregated posterior with the prior. Based on this, Coupled Variational Bayes (CVB)~\cite{dai2018coupled} connects the primal-dual view of ELBO with adversarial learning, where the discriminator takes both data sample and latent code as input. This approach is also adopted in implicit VAE~\cite{fang2019implicit} for text generation. However, the prior used in implicit VAE is still standard Gaussian, while our proposed model uses hyperbolic geometry. 

\begin{figure}[!t]
    \centering
    \includegraphics[width=0.42\textwidth]{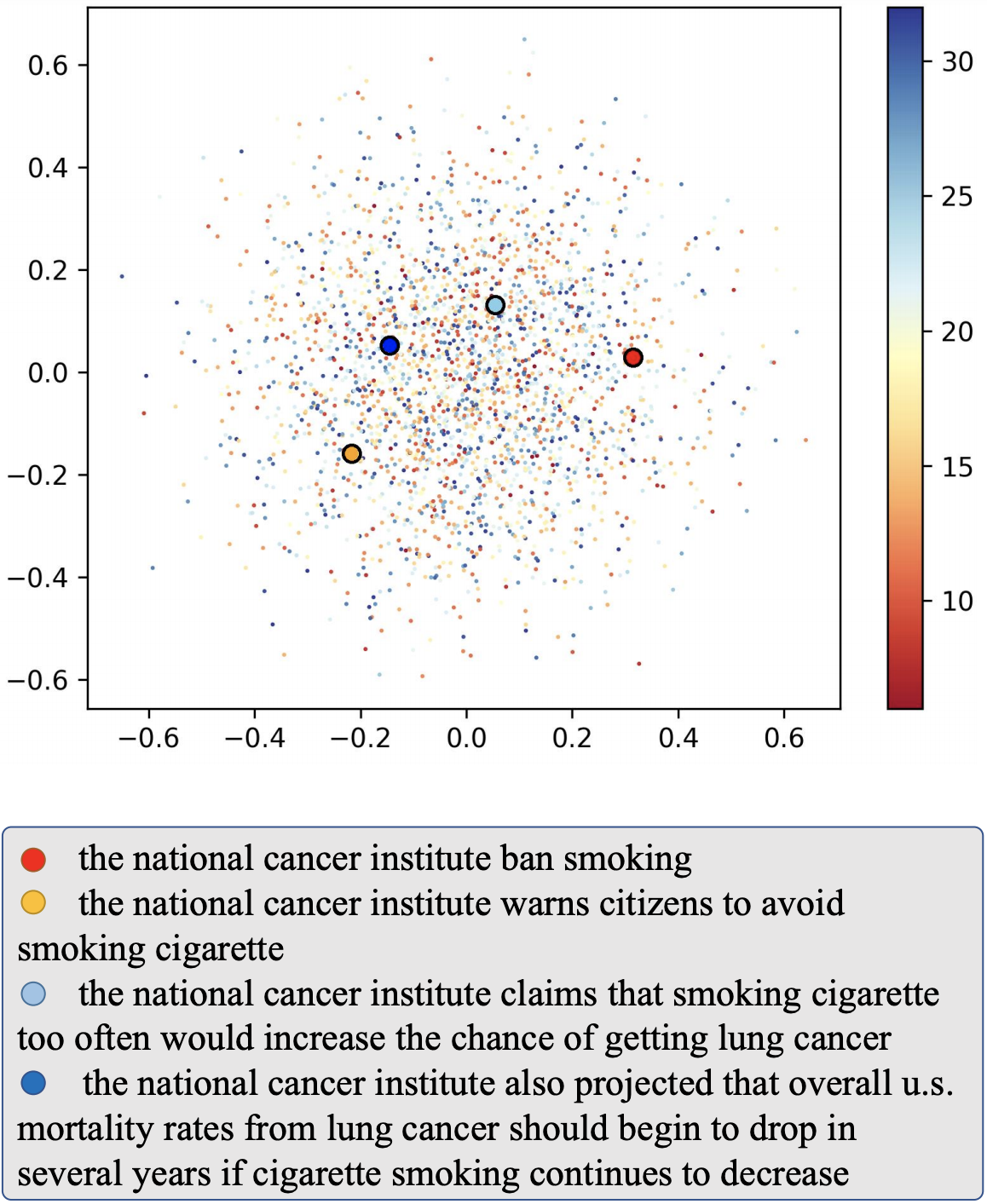}
    
    \caption{Visualization of the Euclidean space (VAE) of 5,000 randomly sampled sentences from the test set of PTB. The four listed sentences are created by modifying data samples.}
    \label{fig:visualization_VAE}
    \vspace{-3mm}
\end{figure}

\section{Additional Details for Experiments}
\subsection{Metrics (Dialogue Response Generation)}
BLEU~\cite{papineni-etal-2002-bleu} is used to measure the amount of $n$-gram overlap between a generated response with the reference. Specifically, BLEU scores of $n<4$ are computed; the average and the maximum scores are considered as $n$-gram precision and $n$-gram recall, respectively. In addition, the BOW embedding metric~\cite{liu2016not} is used to measure cosine similarity between bag-of-word embeddings of response and reference. Three metrics are considered for cosine distance: ($i$) computed by \textbf{greedily} matching words in two utterances; ($ii$) between the \textbf{averaged} embeddings in two utterances; and ($iii$) between the largest \textbf{extreme} values among the embeddings in two utterances. We also follow~\citeauthor{gu2018dialogwae}~and use the \textit{distinct} metric to measure the diversity of generated text. $Dist-n$ is the ratio of unique $n$-grams over all $n$-grams in the generated sentences. \textit{Intra-dist} and \textit{inter-dist} are the average \textit{distinct} values within each generated sample and among all generated samples, respectively.

\begin{figure}[!t]
\vspace{-7mm}
\centering
\begin{subfigure}[b]{0.45\textwidth}
   \includegraphics[width=1\linewidth]{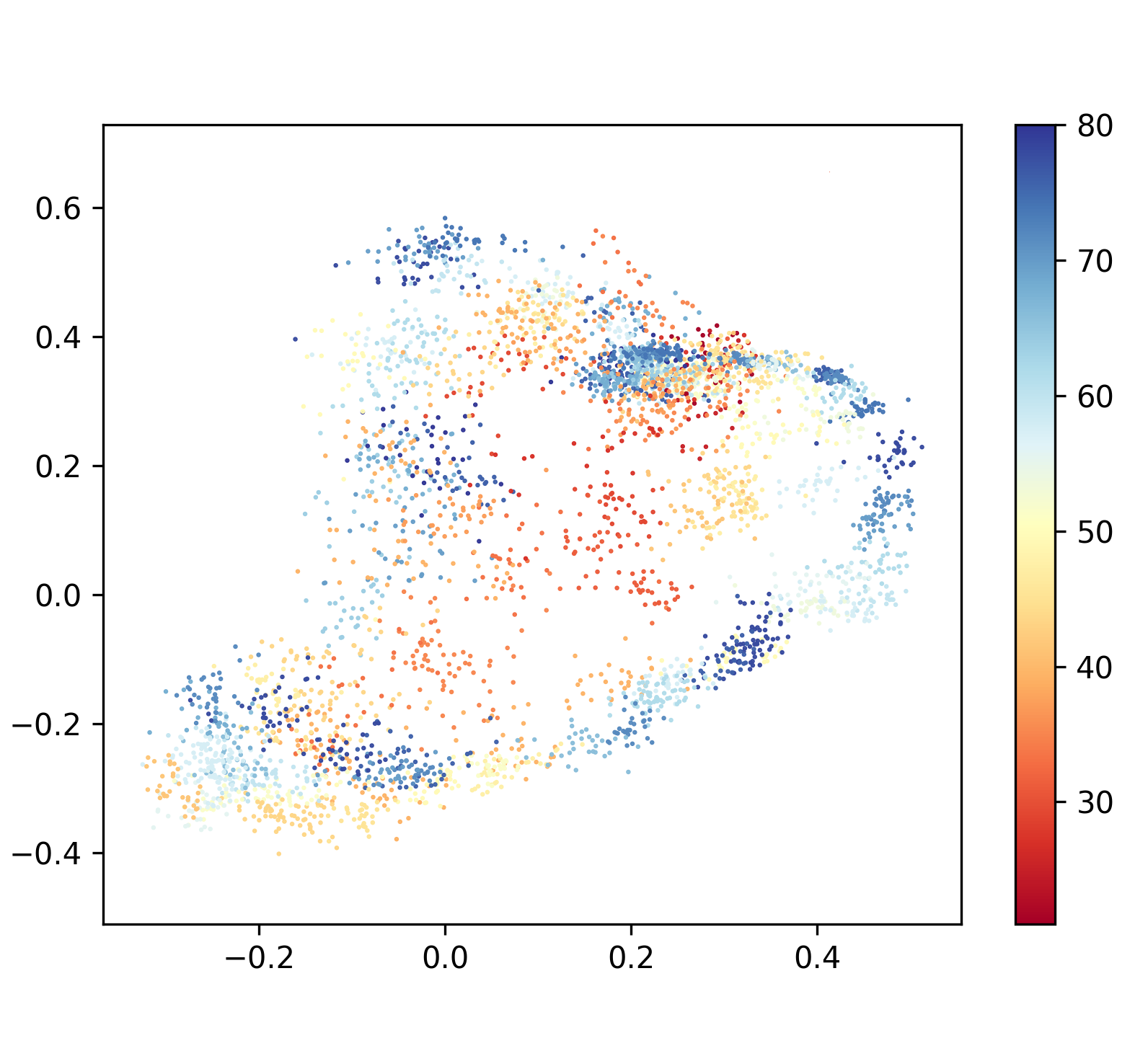}
   \caption{Hyperbolic latent space for Yahoo.}
   \label{fig:sample_yahoo} 
\end{subfigure}
\begin{subfigure}[b]{0.48\textwidth}
   \includegraphics[width=1\linewidth]{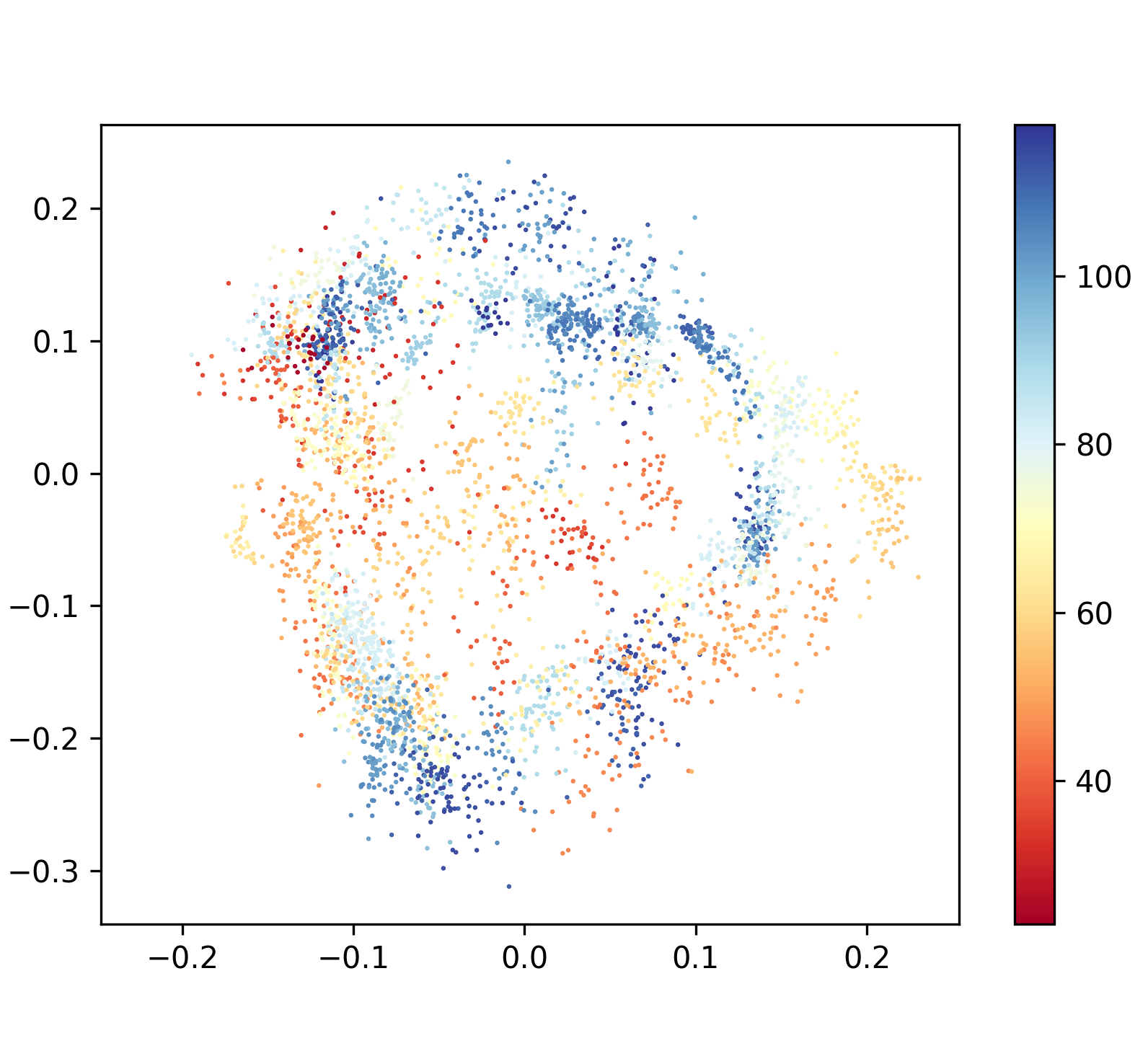}
   \caption{Hyperbolic latent space for Yelp.}
   \label{fig:sample_yelp}
\end{subfigure}
\caption{Visualization of the hyperbolic latent space of 5,000 randomly sampled sentences from different datasets.}
\label{fig:latent_additional}
\vspace{-5mm}
\end{figure}

\begin{table*}[ht]
    \setlength\extrarowheight{5pt} 
    \begin{tabularx}{\textwidth}{C|C}
        \hline
        Input & Sample\\
        \hline
        the national cancer institute ban smoking & the $\langle \text{unk}\rangle$ and drug administration were talking\\
        \hline
        the national cancer institute warns citizens to avoid smoking cigarette & the $\langle \text{unk}\rangle$ and drug administration officials also are used to $\langle \text{unk}\rangle$\\
        \hline
        the national cancer institute claims that smoking cigarette too often would increase the chance of getting lung cancer & the u.s. and drug administration officials say they are n't $\langle \text{unk}\rangle$ to be used by the government \\
        \hline
        the national cancer institute also projected that overall u.s. mortality rates from lung cancer should begin to drop in several years if cigarette smoking continues to decrease & the u.s. and drug administration officials $\langle \text{unk}\rangle$ they are looking for ways to <unk> their own accounts for some of their assets to be sold by some companies\\
        \hline
    \end{tabularx}
    \caption{Generate a sample based on the latent code of a specific input (PTB).}
    \label{tab:language_sample}
\end{table*}

\setlength{\tabcolsep}{9.2pt}
\begin{table*}
    \small
    \centering
    \begin{tabular}{l|ccc|ccc|cc|cc}
    \hline
        \multirow{2}{*}{Model} & \multicolumn{3}{|c|}{BLEU} & \multicolumn{3}{|c|}{BOW} & \multicolumn{2}{|c|}{Intra-dist} & \multicolumn{2}{|c}{Inter-dist}\\
        \cline{2-11}
        & R & P & F1 & A & E & G & dist-1 & dist-2 & dist-1 & dist-2\\
        \hline
        CVAE & 0.265 & 0.222 & 0.242 & 0.923 & 0.543 & 0.811 & 0.938 & 0.973 & 0.177 & 0.222\\
        CVAE-BOW & 0.256 & 0.224 & 0.239 & 0.923 & 0.540 & 0.812 & \textbf{0.947} & 0.976 & 0.165 & 0.206\\
        CVAE-CO & 0.259 & 0.244 & 0.251 & 0.914 & 0.530 & 0.818 & 0.821 & 0.911 & 0.106 & 0.126\\
        DialogWAE & 0.341 & \textbf{0.278} & \textbf{0.306} & 0.948 & 0.578 & 0.846 & 0.830 & 0.940 & 0.327 & 0.583\\
        iVAE & 0.355 & 0.239 & 0.285 & 0.951 & 0.609 & 0.872 & 0.897 & 0.975 & 0.501 & 0.868\\
        \model & \textbf{0.359} & 0.265 & 0.305 & \textbf{0.954} & \textbf{0.616} & \textbf{0.873} & 0.919 & \textbf{0.989} & \textbf{0.511} & \textbf{0.869}\\
    \hline
    \end{tabular}
    \caption{Results on DailyDialog (P: precision, R: recall, A: average, E: extreme, G: greedy). Higher BLEU and BOW Embedding indicate better quality of generated responses. Higher intra/inter-dist means better generation diversity.}
    \label{tab:dailydialog_result}
    \vspace{-3mm}
\end{table*}

\subsection{Additional Implementation Details}
For language modeling, we adopt the LSTM~\cite{hochreiter1997long} for both the encoder and decoder, which have 256 hidden units for PTB, and 1024 hidden units for Yahoo and Yelp. The dimension of the latent code is set to 32. Following~\citet{mathieu2019hierarchical}, the hyper-parameter $c$ is set to $0.7$. We set the vocabulary size to 10,000 for PTB, and 20,000 for both Yahoo and Yelp. 
The word embedding size is 256 for PTB, and 512 for Yahoo and Yelp. 
%
For dialogue response generation, 
we follow~\citet{gu2018dialogwae}, and use the GRU~\cite{cho2014learning} with 300 hidden units in each direction for both the response encoder and context encoder, and 300 hidden units for decoder. The latent code $\bm z$ has a dimension of 200. The size of the vocabulary is set to 10,000, and the word-embedding size is 200, initialized by GloVe~\cite{pennington2014glove}. 

\subsection{Additional Results}
For language modeling, we plot the hyperbolic latent space for Yahoo and Yelp as shown in Figure~\ref{fig:latent_additional}. To demonstrate the hierarchical structure in the generated sentences (\textit{i.e.}, the decoder), we choose 4 sentences (from short to long) with some hierarchy, listed on the left hand side of Table~\ref{tab:language_sample}. These sentences are encoded to hyperbolic space by using a well trained \model. Then, we decode by randomly select a neighbor of each of the 4 latent codes. The output sentences are shown on the right hand side of Table~\ref{tab:language_sample}, demonstrating similar hierarchy as the input sentences. Moreover, we directly measure the generation quality by using PPL and reverse PPL, shown in Table~\ref{tab:ppl_result}. Our APo-VAE achieves consistently better performance.

\setlength{\tabcolsep}{16pt}
\begin{table}[ht]
  \centering
  \begin{tabular}{l|cc}
  \hline
    Model & Forward & Reverse\\
    \hline
    VAE & 18494 & 10149\\
    Cyc-VAE & 3390 & 5587\\
    $\beta$-VAE & 625 & 1897\\
    SA-VAE & 341 & 10651\\
    vMF-VAE & 274 & 2364\\
    iVAE & 116 & 1520\\
    APo-VAE & 109 & 1385\\
  \hline
  \end{tabular}
  \caption{Results on forward and reverse PPL on the PTB dataset.} 
  \label{tab:ppl_result}
  \vspace{-3mm}
\end{table}

\setlength{\tabcolsep}{4.5pt}
\begin{table}[!t]
\small
\centering
\begin{tabular}{l|ccc|ccc}
\hline
\multirow{2}{*}{} & \multicolumn{3}{c|}{vs\, iVAE} & \multicolumn{3}{c}{vs\, DialogWAE} \\
\cline{2-7}
 & win & loss & tie & win & loss & tie \\ 
\hline
Informativeness & 45.4 & 26.9 & 17.7 & 46.1 & 26.5 & 27.4 \\
Coherence & 40.1 & 25.9 & 24.0 & 40.7 & 24.2 & 25.1 \\
Diversity & 43.9 & 30.8 & 25.3 & 47.5 & 31.4 & 21.1 \\
\hline
\end{tabular}
\caption{Human evaluation results on DailyDialog. Win/loss/tie indicates the percentage of responses generated by APo-VAE being better/worse/equal to the compared model.}
\label{tab:human_eval_2}
\vspace{-3mm}
\end{table}


For dialog response generation, we include additional results on the DailyDialog dataset~\cite{li2017dailydialog}, which contains 13k daily conversations for an English learner in daily life. We also provide examples of generated responses along with their corresponding dialog context in Table~\ref{tab:dialogue_sample}. Samples generated by \model~are more relevant to the contexts than the baseline models. In addition, \model~is capable of providing both positive and negative responses, demonstrating better generation diversity. The human evalutation results for DailyDialog can be found in Table~\ref{tab:human_eval_2}.

\subsection{Human Evaluation}
\begin{figure*}[ht]
    \centering
    \includegraphics[width=0.99\textwidth]{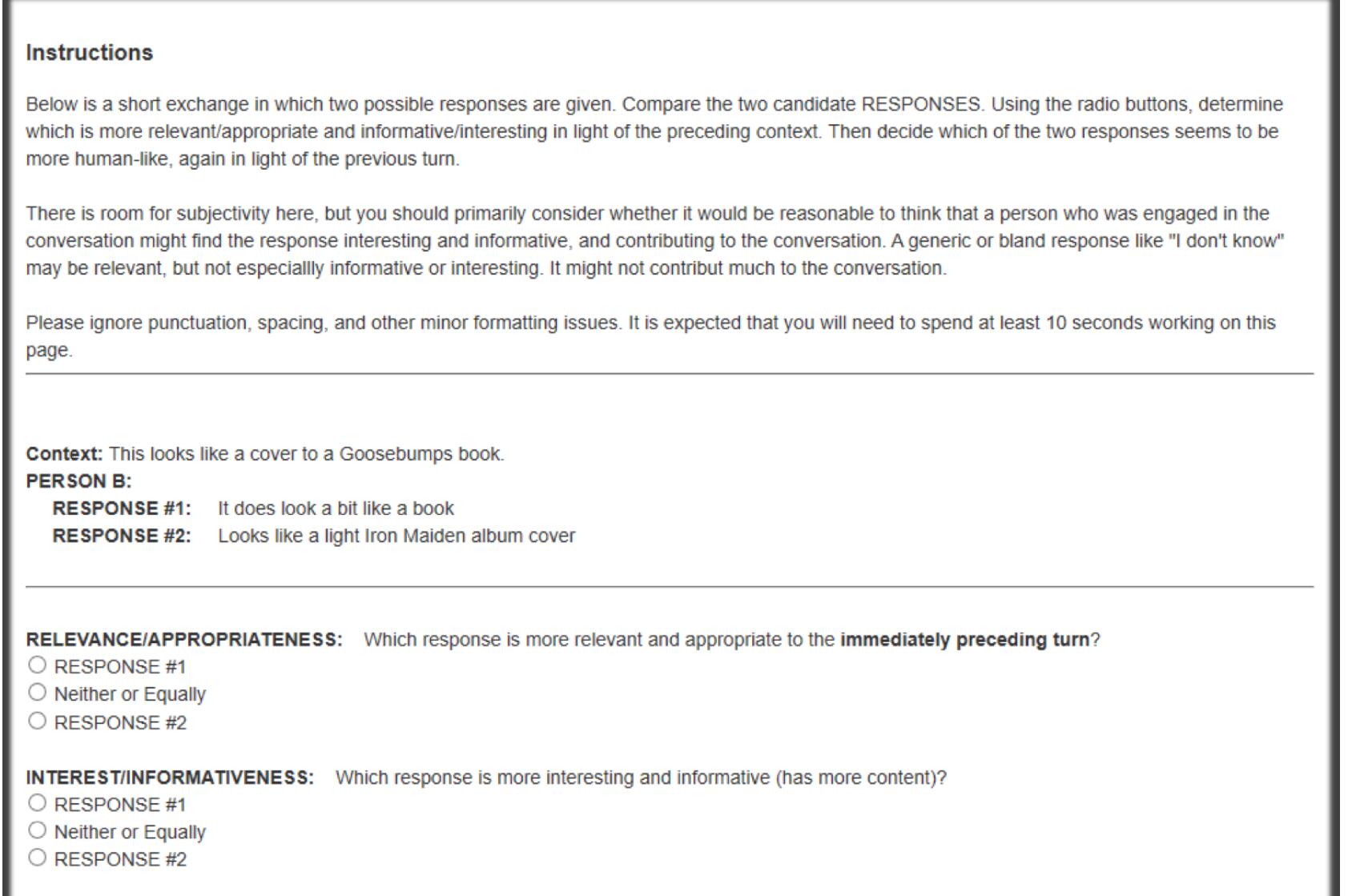}
    \caption{AMT interface for user study. Human judges are required to compare responses given context text.}
    \label{fig:human_eval}
\end{figure*}

We provide the instruction of human evaluation on the dialog response generation task in Figure~\ref{fig:human_eval}.

\setlength{\tabcolsep}{14pt}
\begin{table*}[ht]

    \begin{tabularx}{\textwidth}{p{3cm}|p{3cm}|p{3cm}|p{3cm}}
        \hline
        Context & DialogWAE & iVAE & \model\\
        \hline
        i would wondering if you like to stay here for a few days? & Eg.1: ok. & Eg.1: I'm not going to take a look. & Eg.1: good idea. how about shopping tomorrow morning? \\
        \cline{2-4}
         & Eg.2: great, thanks for asking. & Eg.2: is it a problem? & Eg.2: of course. would you like to have lunch with me? \\
        \cline{2-4}
         & Eg.3: next week is a good idea. & Eg.3: it is a day vacation for you here in the us. & Eg.3: i'm afraid i can't make it to the next few days. \\
        \hline
        i have to check out today. i'd like my bill ready by 10. & Eg.1: that's a long time. & Eg.1: that's the problem, sir? & Eg.1: i'm sorry to hear that. are you looking for another one?\\
        \cline{2-4}
         & Eg.2: yes. & Eg.2: i was the only child in the room. & Eg.2: no problem. may i have it now?\\
        \cline{2-4}
         & Eg.3: i'm afraid you can't do that. & Eg.3: you can get me a little time, and the red will take you get off. & Eg.3: i see, sir. what is the room number?\\
        \hline
        the weather has been terrible this week. & Eg.1: i'm sure we can do it again. & Eg.1: i've got a few things i'll get you there is a real one. & Eg.1:  yes. i have been looking forward to better days. \\
        \cline{2-4}
         & Eg.2: oh, this is the correct. & Eg.2: what is the rest of the days. & Eg.2: i don't think it is the best time for us to visit.\\
        \cline{2-4}
         & Eg.3: what kind of job do you like to see? & Eg.3: oh, no. & Eg.3: yes, we did. \\
        \hline
        i wonder whether i could possibly borrow your new bicycle now. & Eg.1: that's a great idea. what did you do? & Eg.1: a problem. what do you think of this one? & Eg.1: of course. let’s see. \\
        \cline{2-4}
         & Eg.2: i'd like to tell you a fees, but i'm not sure i can do that. & Eg.2: yes sure, thanks. you are very tough.  & Eg.2: i sorry to hear that. anything else? do you have any other account?\\
        \cline{2-4}
         & Eg.3: oh, what is your problem? & Eg.3: what is the problem? what time? oh yes, it's been a long time. let me see. & Eg.3: i would love to do that. there have been more 1 for you. \\
        \hline
        
    \end{tabularx}
    \caption{Examples of context-response pairs for different models. `Eg.i' means the $i$-th sample.}
    \label{tab:dialogue_sample}
\end{table*}

\end{document}